\documentclass[10pt,twocolumn,letterpaper]{article}

\usepackage[pagenumbers]{iccv} 
\usepackage{multirow}
\usepackage{makecell}

\definecolor{iccvblue}{rgb}{0.21,0.49,0.74}
\usepackage[pagebackref,breaklinks,colorlinks,allcolors=iccvblue]{hyperref}
\usepackage{array}
\definecolor{mygray}{gray}{.9}

\usepackage{multirow}
\usepackage{tcolorbox}
\newcommand\blfootnote[1]{%
  \begingroup
  \renewcommand\thefootnote{}\footnote{#1}%
  \addtocounter{footnote}{-1}%
  \endgroup
}
\usepackage{marvosym}
\def\paperID{14075} 
\def\confName{ICCV}
\def\confYear{2025}

\title{\textit{SpaceVLLM}: Endowing Multimodal Large Language Model with Spatio-Temporal Video Grounding Capability}

\author{
Jiankang Wang$^{1}$\footnotemark[1] \quad
Zhihan Zhang$^{1}$\footnotemark[1] \quad
Zhihang Liu$^{1}$ \quad
Yang Li$^{2}$ \quad
Jiannan Ge$^{1}$  \quad \\
Hongtao Xie$^{1}$\textsuperscript{\Letter}  \quad
Yongdong Zhang$^{1}$  \quad \\
$^{1}$University of Science and Technology of China \\
$^{2}$Renmin University of China \\
{\tt\small \{wangjiankang, zhangzhihan, liuzhihang, gejn\}@mail.ustc.edu.cn, liyang03@ruc.edu.cn}\\
{\tt\small \{htxie, zhyd73\}@ustc.edu.cn}
\vspace{-2mm}
}

\begin{document}
\maketitle
\blfootnote{$^*$ Equal contribution.\ \ \textsuperscript{\Letter} Corresponding author.}

\begin{abstract}
Multimodal large language models (MLLMs) have made remarkable progress in either temporal or spatial localization. However, they struggle to perform spatio-temporal video grounding. This limitation stems from two major challenges. Firstly, it is difficult to extract accurate spatio-temporal information of each frame in the video. Secondly, the substantial number of visual tokens makes it challenging to precisely map visual tokens of each frame to their corresponding spatial coordinates. To address these issues, we introduce \textit{SpaceVLLM}, a MLLM endowed with spatio-temporal video grounding capability. Specifically, we adopt a set of interleaved Spatio-Temporal Aware Queries to capture temporal perception and dynamic spatial information. Moreover, we propose a Query-Guided Space Decoder to establish a corresponding connection between the queries and spatial coordinates. Additionally, due to the lack of spatio-temporal datasets, we construct the Unified Spatio-Temporal Grounding (Uni-STG) dataset, comprising 480K instances across three tasks. This dataset fully exploits the potential of MLLM to simultaneously facilitate localization in both temporal and spatial dimensions. Extensive experiments demonstrate that \textit{SpaceVLLM} achieves the state-of-the-art performance across 11 benchmarks covering temporal, spatial, spatio-temporal and video understanding tasks, highlighting the effectiveness of our approach. Our code, datasets and model will be released at \url{https://github.com/Jayce1kk/SpaceVLLM}.
\end{abstract}
    
\section{Introduction}
\label{sec:intro}

\begin{figure}[t]
\centering
\includegraphics[width=\linewidth]{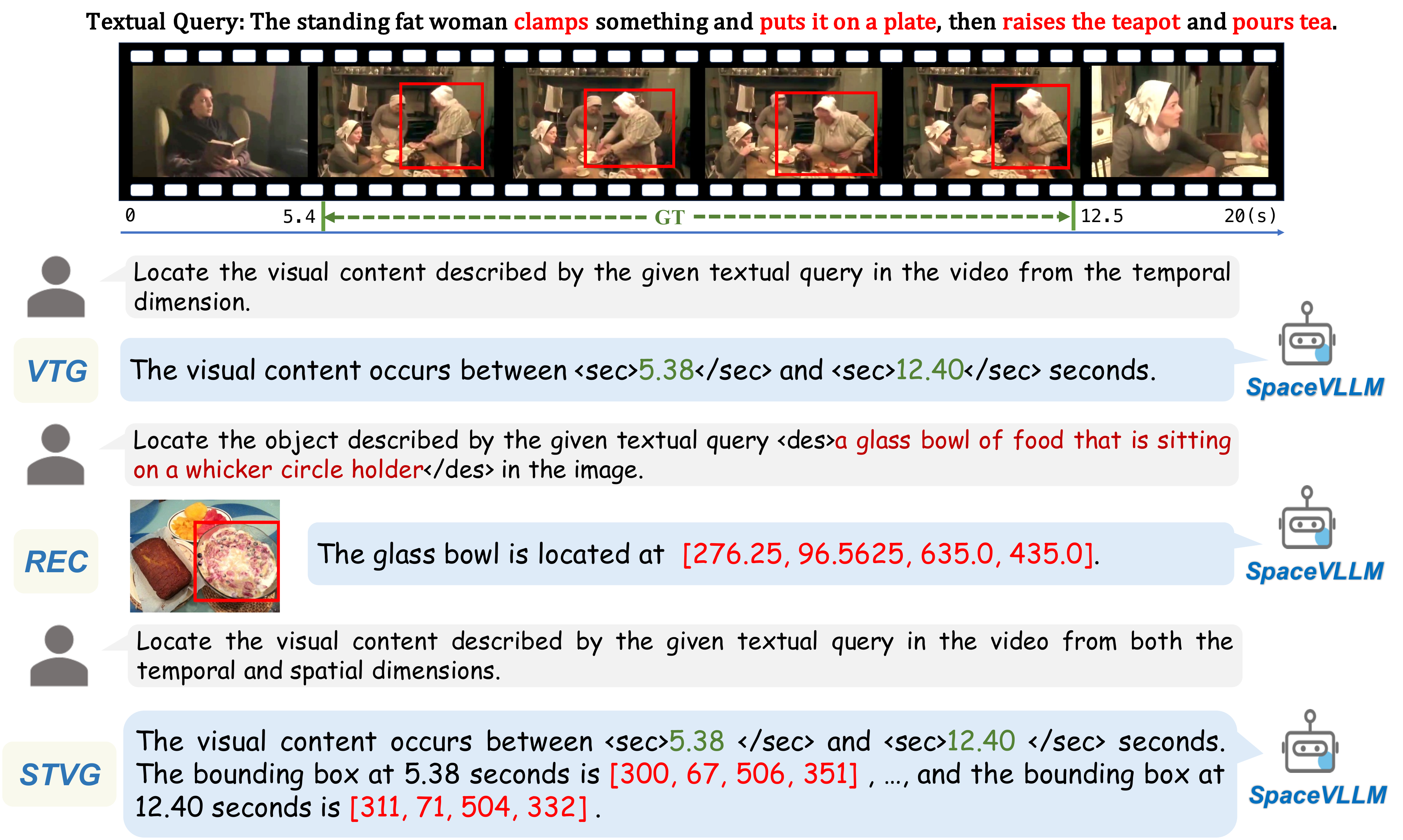}
\caption{Example of the Video Temporal Grounding (VTG), Referring Expression Comprehension (REC) and Spatio-Temporal Video Grounding (STVG) tasks in the proposed \textit{SpaceVLLM}. }
\label{fig:intro}
\vspace{-10pt}
\end{figure}

Multimodal large language models (MLLMs) have recently demonstrated significant advancements in multimodal understanding~\cite{ref:videochat,ref:videollava,ref:llavaov,ref:llavavideo}. With the rapid development of MLLMs, an increasing number of works focus on multimodal comprehension from either a temporal or spatial perspective. Some studies~\cite{ref:grounded, ref:vtgllm, ref:vtimellm,ref:trace,ref:timechat} aim to enhance the ability of MLLMs to perceive temporal information in tasks such as Video Temporal Grounding, while others~\cite{ref:shikra,ref:elysium,ref:groma,ref:fer,ref:groundinggpt} concentrate on localizing referred objects within a single image. 
However, the challenging task of Spatio-Temporal Video Grounding (STVG)~\cite{ref:vidstvg}, which requires locating a spatio-temporal tube corresponding to a specific instance described by a caption, remains underexplored by existing MLLMs. 

How can MLLMs be endowed with Spatio-Temporal Video Grounding capability? One straightforward thought is to use MLLMs~\cite{ref:groundinggpt} equipped with video temporal grounding and image localization capabilities to determine the start and end timestamps first, and then process each frame within this range to perform image localization. Nevertheless, when the given caption describes a person's action, this approach often fails to extract accurate spatio-temporal information, as static images lack temporal awareness and the dynamic spatial details necessary for precise localization. Another idea is to allow powerful MLLMs~\cite{ref:qwen2.5,ref:llavavideo} to generate the time range and the coordinates of each frame in one step through instruction tuning. However, a video inherently contains a vast number of visual tokens, and MLLMs are required to output considerable coordinates at once. This makes it difficult to accurately associate each spatial coordinate with the corresponding frame's visual tokens.

To address the above challenges, we propose \textit{SpaceVLLM}, the first MLLM capable of simultaneously integrating spatial and temporal information to perform the Spatio-Temporal Video Grounding (STVG) task. To acquire precise spatio-temporal details of each frame in the video, we pair each video frame with a Spatio-Temporal Aware Query. These queries are interleaved with the visual features of the video frames to capture both static visual information and inter-frame dynamic spatial cues. Furthermore, the ordered positions of the queries inherently make them time-sensitive. To map the extracted information into coordinates, we propose a Query-Guided Space Decoder which connects the query of the corresponding frame with the spatial coordinate. Specifically, a dual cross attention module is first applied to enhance the spatial information for the queries. Then we employ a lightweight space head to generate the accurate coordinates. \textit{SpaceVLLM} significantly enhances the MLLM’s ability to comprehend videos across multiple dimensions. Figure \ref{fig:intro} shows our model's temporal, spatial and spatio-temporal localization capabilities.

Additionally, most existing datasets are designed to address either temporal grounding~\cite{ref:didemo,ref:vtgllm,ref:intervid} or spatial grounding~\cite{ref:visual,ref:refcoco,ref:refcocog} in isolation. The lack of spatio-temporal datasets limits the model’s ability to fully capture the fine-grained details of a video across both temporal and spatial domains.
To tackle this issue, we design a pipeline to synthesize a comprehensive spatio-temporal video grounding dataset, characterized by high-quality annotations and diverse object categories. Then we construct the Unified Spatio-Temporal Grounding dataset (Uni-STG), which encompasses three tasks: Video Temporal Grounding (VTG), Referring Expression Comprehension (REC) and Spatio-Temporal Video Grounding (STVG). In total, the dataset contains 480K samples, facilitating fine-grained spatio-temporal understanding. To ensure that our model retains its general understanding capabilities, we incorporate multiple types of datasets for multi-task instruction tuning. Extensive experiments demonstrate that our model achieves state-of-the-art performance on 11 benchmarks, including temporal, spatial, spatio-temporal, and video understanding, emphasizing the effectiveness of our model.

Our contributions can be summarized as follows:
\begin{itemize}  
    \item We propose \textit{SpaceVLLM}, the first MLLM equipped with spatio-temporal video grounding capability. To achieve this, we utilize the Spatio-Temporal Aware Queries to accurately extract spatio-temporal information and introduce Query-Guided Spatial Decoder to precisely map these queries to their corresponding spatial coordinates.
    
    \item We propose a Unified Spatio-Temporal Grounding dataset (Uni-STG) with 3 tasks and 480K instances, facilitating fine-grained spatial-temporal understanding.
    
    \item We conduct experiments on 11 benchmarks, including temporal, spatial, spatio-temporal and understanding tasks, achieving the state-of-the-art performance.
\end{itemize}
\section{Related Work}
\label{sec:formatting}

\subsection{Spatial-Temporal Video Grounding}
Spatial-Temporal Video Grounding aims to localize the target object temporally and spatially according to a language query. Early methods~\cite{ref:hcstvg, ref:object, ref:vidstvg} adopt a two-stage paradigm, which first utilizes a pretrained detector like Faster-RCNN~\cite{ref:fasterrcnn} to obtain the candidate region proposals and then find the correct region proposal. However, these methods are restricted by the ability of the pre-trained detectors. Recent methods~\cite{ref:embrace, ref:tubedetr, ref:stvgbert} follow a one-stage paradigm to directly generate spatio-temporal object proposals without applying the pre-trained object detectors. STVGBert~\cite{ref:stvgbert} devises a ST-ViLBert module to generate bounding boxes and predict the start and end frames to produce the predicted object tube. TubeDETR~\cite{ref:tubedetr} utilizes a video-text encoder along with a spatio-temporal transformer decoder to localize in two dimensions. The method of~\cite{ref:coll} designs a static and a dynamic vision-language stream to collaboratively reason the target tube. Currently proposed CG-STVG~\cite{ref:cgstvg} mines the instance visual context from the video to guide target localization. 
In this paper, we first employ MLLM to empower the potential to localize the target object both temporally and spatially.

\begin{figure*}[t]
\centering
\includegraphics[width=1\linewidth]{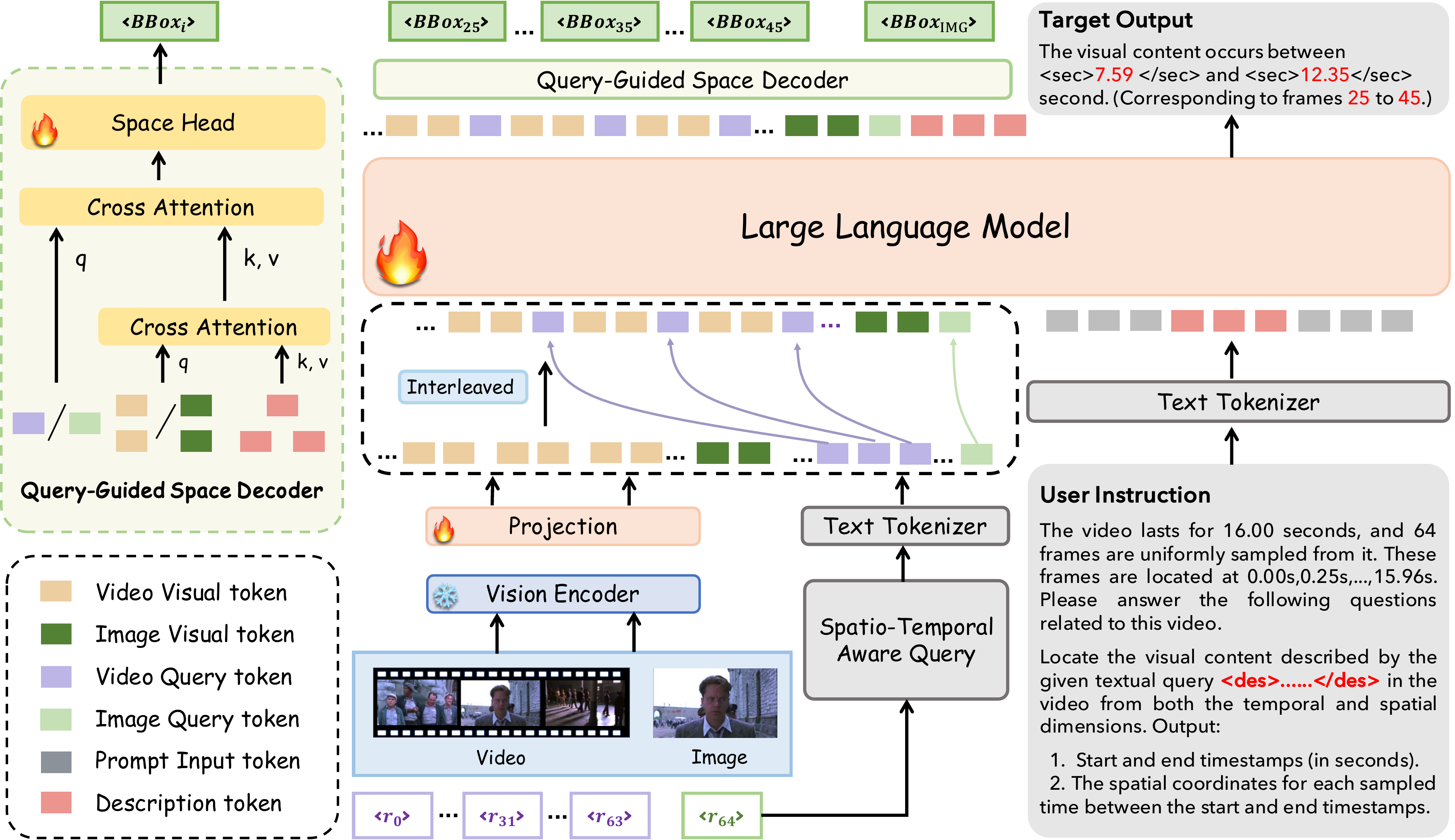}
\vspace{-4mm}
\caption{The Overall Architecture of \textit{SpaceVLLM}. In \textit{SpaceVLLM}, A set of ordered Spatio-Temporal Aware Queries is interleaved with visual tokens of each video frame to capture spatio-temporal information. The LLM's last-layer query embeddings, combined with corresponding visual and description embeddings, are fed into the Query-Guided Space Decoder to predict frame-wise coordinates. }
\label{fig:framework}
\vspace{-4mm}
\end{figure*}

\subsection{Multimodal Large Language Models}
Recently Multimodal Large Language Models (MLLMs) have shown significant progress in understanding videos. Traditional video LLMs have achieved remarkable performance in the task of visual question answering, video caption and reasoning~\cite{ref:videochat,ref:videollama,ref:videochatgpt,ref:gemini,ref:videollava}. On the one hand, some works~\cite{ref:vtgllm, ref:vtimellm, ref:grounded} explore the potential of locating the start and end timestamps of the video. VTimeLLM~\cite{ref:vtimellm} adopts a boundary-aware three-stage training strategy to empower the LLM to grasp video moments. Recently TRACE~\cite{ref:trace} proposes a causal event modeling framework to pinpoint the timestamps. On the other hand, some studies~\cite{ref:shikra,ref:llavagrounding, ref:groma} are adept at understanding the referring expressions and locating the spatial coordinate in the image. Groma~\cite{ref:groma} designs a localized visual tokenization mechanism for fine-grained region captioning and visual grounding. GroundingGPT~\cite{ref:groundinggpt} proposes a language enhanced multimodal grounding model to pinpoint the timestamps in the video and locate the referring object in the image. However, existing models struggle to perform spatial-temporal grounding, as they cannot simultaneously capture both temporal and spatial information. In this paper, we propose \textit{SpaceVLLM}, a novel model that empowers the MLLM with joint spatial and temporal capabilities.

\section{Method}
\label{sec:finalcopy}
In this section, we first illustrate our model \textit{SpaceVLLM} in \ref{sec:model}. Next, we introduce our Unified Spatio-Temporal Grounding (Uni-STG) dataset and model training in \ref{sec:dataset}. 
\vspace{-0.25mm}
\subsection{Architecture}
\label{sec:model}
\subsubsection{Overview}
\label{sec:overview}
The overall architecture of \textit{SpaceVLLM} is illustrated in Figure \ref{fig:framework}. Images or videos are processed through a vision encoder to extract features. Each frame is paired with a Spatio-Temporal Aware Query, while the static image is paired with a specific spatial-aware query. Next, these visual tokens and queries are concatenated with descriptions and user instructions, which are then fed into the LLM to generate responses. Finally, the LLM last-layer embedding of queries along with the visual embedding and text embedding is processed through our Query-Guided Space Decoder to generate frame-wise coordinates.

\subsubsection{Spatio-Temporal Aware Query}
\label{sec:query}

To leverage rich cues from the videos, we define a set of special tokens as spatio-temporal aware queries. In specific, we first sample a set of frames $\mathbf{X_v} = \{x_i\}_{i=0}^{N_{v}-1}$ of length $N_{v}$ from the video. $N_{v}$ is set to 1 for the image input. We then construct $N_{v}+1$ learnable tokens $\mathbf{R} = \{\texttt{<}r_i\texttt{>}\}_{i=0}^{N_v}$. The first $N_{v}$ tokens $\{\texttt{<}r_i\texttt{>}\}_{i=0}^{N_v-1}$ serve as spatio-temporal aware queries for video grounding, and the terminal token $\texttt{<}r_{N_v}\texttt{>}$ operates as a spatial-aware query for image grounding. 

The vision encoder $\mathcal{G}(\cdot)$ extracts visual features from each frame, which are then mapped into the embedding space by a projector $\mathcal{P}(\cdot)$, yielding a sequence of visual embeddings $\mathbf{V}$. Simultaneously, the text tokenizer $\mathcal{F}_{txt}(\cdot)$ encodes the specific tokens into their word embeddings $\mathbf{Q}$:
\begin{equation}
\begin{aligned}
\mathbf{V} &= \mathcal{P}(\mathcal{G}(\mathbf{X_v})), \quad \mathbf{V} \in \mathbb{R}^{N_{v} \times S \times D}, \\
\mathbf{Q} &= \mathcal{F}_{txt}(\mathbf{R}), \quad \mathbf{Q} \in \mathbb{R}^{N_v \times 1 \times D},
\end{aligned}
\end{equation}
where $N_v$ refers to the number of the sampled frames and $S$ represents the number of the visual tokens for each frame. Finally, the spatio-temporal enhanced visual representation $\mathbf{H}$ is constructed through interleaved concatenation of visual and query embeddings:
\vspace{-2mm}
\begin{equation}
\mathbf{H} = \bigoplus_{i=0}^{N_v-1} \left( \mathbf{v}_i \oplus \mathbf{q}_i \right) , \mathbf{H}\in \mathbb{R}^{N_v \times (S+1) \times D},
\end{equation}
where $v_i\in \mathbf{V}$ denotes the $i$-th frame's visual features, $q_i\in \mathbf{Q}$ represents corresponding spatio-temporal aware query features, and $\oplus$ indicates row-wise concatenation. Finally, $\mathbf{H}$, along with the user instruction, is fed into the LLM to obtain the text output $\hat{\mathbf{y}}_{txt}$.
Each spatio-temporal aware query is inserted between the visual tokens of two adjacent frames, which is expected to learn spatial details from the corresponding frame and dynamic information between the adjacent frames. Due to their sequential positions, each query also retains temporal information.
Through the LLM, these queries are rich in temporal awareness and capture inter-frame dynamic spatial information precisely.

To make the sampled frame sensitive to the time, we add time instructions like ``\texttt{The video lasts for 20 seconds, and 64 frames are uniformly sampled from it. These frames are located at 0.00s,0.28s,...19.96s.}" Additionally, we adjust the time boundary according to the sampled frame's seconds. This way can help the large language model better perceive the time.

\subsubsection{Query-Guided Space Decoder}
After obtaining the queries enriched with spatio-temporal information, we propose a Query-Guided Space Decoder to map these queries into coordinates, including Dual Cross Attention and Space Head.

\label{sec:head}
\noindent\textbf{Dual Cross Attention.} To connect the LLM's output with the box coordinates, we devise a dual cross attention module to enhance the spatial information of the current frame. Specifically, cross attention between visual embeddings and textual embeddings of the caption is first employed to obtain text-enhanced visual features. Next, the spatio-temporal aware queries and enhanced visual embedding calculate cross attention to strengthen the spatial connections between queries and visual representations, which can be formulated as:
\vspace{-2mm}
\begin{equation}
\begin{aligned}
\mathbf{V'}_{eh_i} = Attention(\mathbf{Q'}_{vi},\mathbf{K}_t,\mathbf{V}_t),  \\
\mathbf{S}_i = Attention( \mathbf{Q'}_{si},\mathbf{V'}_{eh_i},\mathbf{V'}_{eh_i}),
\end{aligned}
\end{equation}
where $\mathbf{Q'}_{vi}$ and $\mathbf{Q'}_{si}$ represent the LLM last-layer embeddings of the visual tokens from the $i$-th frame and the corresponding spatio-temporal aware query, respectively. $\mathbf{K}_t$ and $\mathbf{V}_t$ represents the LLM last-layer embedding of the input caption. $\mathbf{V'}_{eh_i}$ represents the text-enhanced visual tokens of the $i$-th frame. $\mathbf{S}_i$ represents the spatial information of the $i$-th frame related to the caption. Note that we do not introduce additional parameters for training through the dual cross attention.

\noindent\textbf{Space Head.} To obtain the representation of the box coordinate of the $i$-th frame, we pass the $\mathbf{S}_i$ through a Multi-Layer Percepton (MLP):
\vspace{-2mm}
\begin{equation}
\mathbf{b}_i(c_x,c_y,w,h)= \mathrm{MLP}(\mathbf{S}_i),
\vspace{-2mm}
\end{equation}
where the four-dimensional coordinates refer the center point $c_x$, center point $c_y$, width $w$ and height $h$ of the box. Through an effective MLP, we can get accurate coordinates.

\subsubsection{Training Objectives}

For every spatio-temporal training sample, we decompose the loss into two components: the time loss $\mathcal{L}_{time}$ and the space loss $\mathcal{L}_{space}$. Each sample has a ground-truth bounding box sequence $\mathbf{B} = \{b_t\}_{t=t_s}^{t_e}$ and the corresponding text containing the start and end timestamps $\mathbf{y}_{txt}$. For spatial localization, we involve the box prediction loss $\mathcal{L}_{space}$ with corresponding loss weights $\lambda_{L_1}$ and  $\lambda_{giou}$ as follows:
\begin{equation}
    \mathcal{L}_{space} = \lambda_{L_1} \mathcal{L}_{L_1}(\hat{B},B) + \lambda_{giou} \mathcal{L}_{giou}(\hat{B},B),
\end{equation}
where $\mathcal{L}_{L_1}$ and $\mathcal{L}_{\text{giou}}$ are the $L_1$ loss and generalized IoU loss~\cite{ref:iou} on the bounding boxes respectively. Note that $\mathcal{L}_{space}$ only considers predictions in $[t_s,t_e]$. As for temporal localization, we leverage MLLM to predict the time range. The time loss is computed using the auto-regressive cross-entropy loss for text generation. Given the ground-truth targets $\mathbf{y}_{txt}$, $\mathcal{L}_{time}$ can be denoted as  $\mathcal{L}_{time}=\mathcal{L}_{txt}(\hat{\mathbf{y}}_{txt}, \mathbf{y}_{txt})$, where $\hat{\mathbf{y}}_{txt}$ refers the LLM's text output. The overall objective $\mathcal{L}$ is the weighted sum of these losses, determined by $\lambda_{time}$ and $\lambda_{space}$:
\begin{equation}
\mathcal{L}=\lambda_{time}\mathcal{L}_{time}+\lambda_{space}\mathcal{L}_{space}.
\end{equation}

\subsubsection{Generation of Spatial-Temporal Tube}
Given a video and a user's instruction, \textit{SpaceVLLM} will output the precise time range in text form. After converting the time range to the frame range $[f_s, f_e]$, we extract the visual features of the frames within the range $\mathbf{V'} = \{\mathbf{V'}_i\}_{i=f_s}^{f_e}$ and the corresponding spatio-temporal aware queries $\mathbf{Q'} = \{\mathbf{Q'}_i\}_{i=f_s}^{f_e}$. Both, along with the textual features, are fed into the query-guided space decoder to obtain the predicted boxes $\mathbf{B} = \{b_i\}_{i=f_s}^{f_e}$.
When predicting the temporal range and spatial coordinates, the output for each frame's coordinate corresponds precisely to its spatio-temporal information. As a result, our method achieves superior accuracy, better interpretability, and faster inference speed.

\begin{table*}[t]
\centering
    \renewcommand{\arraystretch}{1.05}
     \resizebox{\textwidth}{!}{
     \begin{tabular}{llll}
         \toprule
         \textbf{Stage} & \textbf{Task}  & \textbf{Dataset} & \textbf{Samples} \\

         \midrule
         \multirow{6}{*}{\makecell[c]{Multi-Task \\ Instruction Tuning}}
         & Video Temporal Grounding  &  DiDeMo~\cite{ref:didemo}, Charades-STA~\cite{ref:charades}, TACoS~\cite{ref:tacos} &50K\\
         & Spatio-Temporal Video Grounding  & Synthesized Data & 110K\\
         & Referring Expression Comprehension & RefCOCO~\cite{ref:refcoco}, RefCOCO+~\cite{ref:refcoco}, RefCOCOg~\cite{ref:refcocog} & 320K \\
         & Visual Question Answering  & NeXTQA~\cite{ref:nextqa}, ActivityNetQA~\cite{ref:activitynetqa}, CLEVRER~\cite{ref:clevrer} & 100K \\
         & Video Caption & ShareGemini~\cite{ref:sharegemini}, ShareGPT4Video~\cite{ref:chen2024sharegpt4video} & 50K\\
         & Conversation  & VCG-Plus~\cite{ref:vcgplus} & 50K \\
         \bottomrule
    \end{tabular}}
     \caption{Overview of Datasets Used in Training for Various Tasks.}
\label{tab:dataset}
\vspace{-2mm}
\end{table*}

\subsection{Training Dataset and Model Training}
\label{sec:dataset}
\subsubsection{Data Synthesis}
\label{sec:datasyn}
\begin{figure}[t]
\centering
\includegraphics[width=\linewidth]{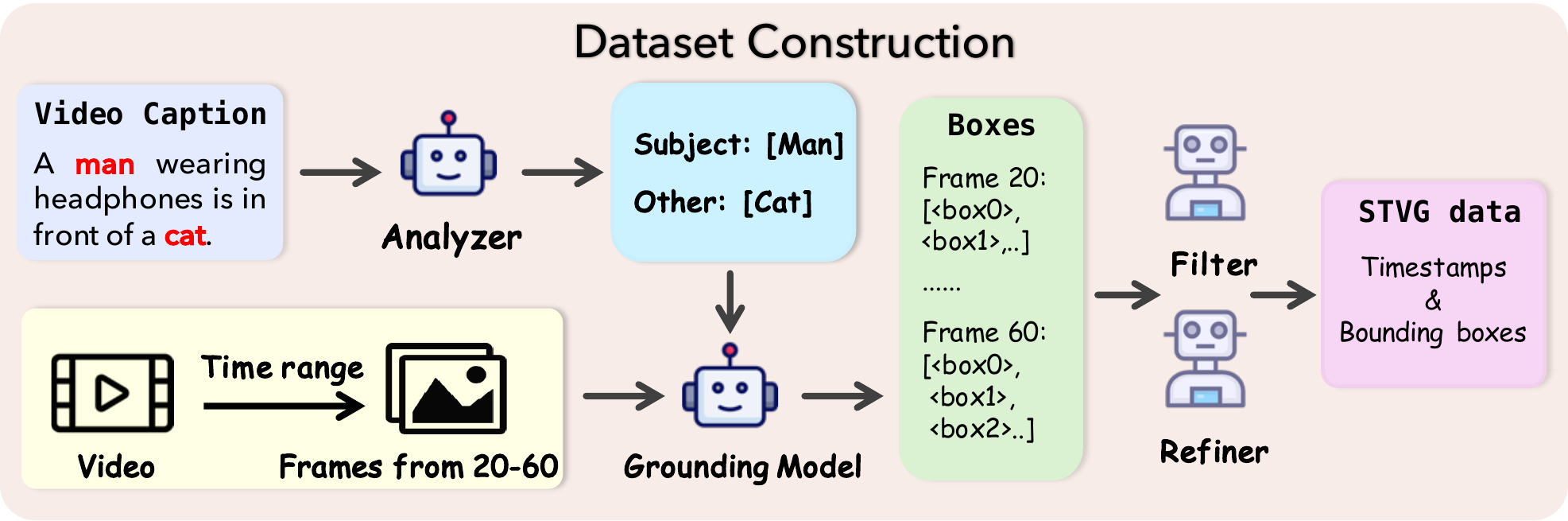}
\caption{Pipeline of data synthesis for STVG task.}
\label{fig:dataset1}
\vspace{-10pt}
\end{figure}

\begin{figure}[t]
\centering
\includegraphics[width=\linewidth]{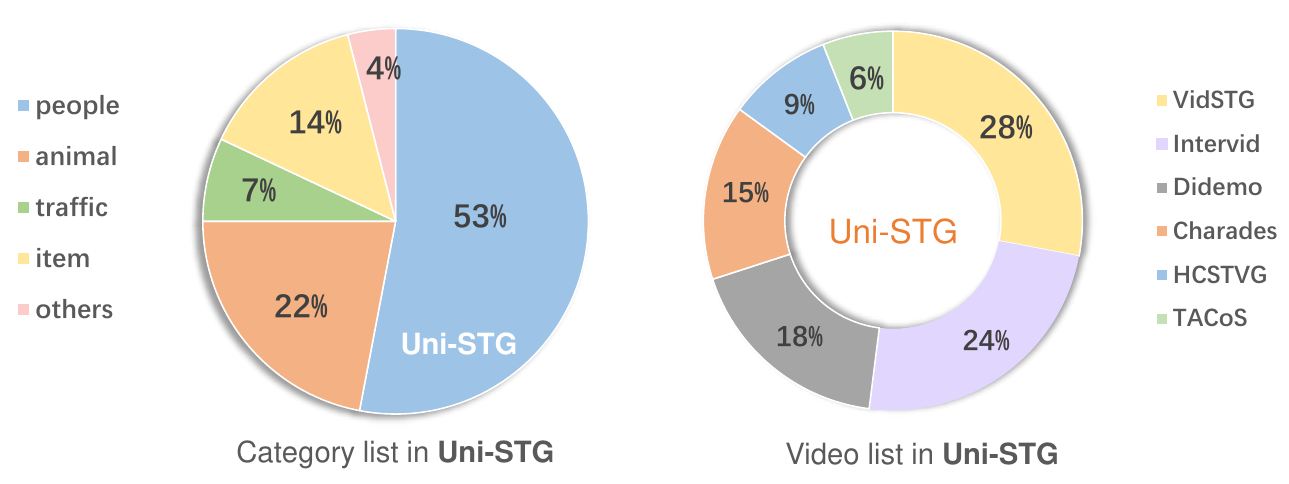}
\caption{Data characteristics of Uni-STG for STVG task.}
\label{fig:dataset2}
\vspace{-10pt}
\end{figure}

To enhance the fine-grained temporal-spatial understanding of LLM, we construct a Unified Spatio-Temporal Grounding (Uni-STG) dataset comprising 480K instances. It contains three tasks, including Video Temporal Grounding (VTG), Referring Expression Comprehension (REC) and Spatio-Temporal Video Grounding (STVG). For the first two tasks, we collect existing datasets and devise task instructions and output formats for each task. For the Spatio-Temporal Video Grounding, due to the limited existing dataset for understanding video both temporally and spatially, we first collect a wide range of video data from Charades-STA~\cite{ref:charades}, TACoS~\cite{ref:tacos}, DiDeMo~\cite{ref:didemo} and Intervid~\cite{ref:intervid}. Their videos feature a wide range of objects, including people, animals, items, and more. Figure \ref{fig:dataset2} presents the composition of our Unified Spatio-Temporal Grounding (Uni-STG) dataset for the STVG task, showcasing the diversity of video sources and object categories. Then we synthesize the Spatio-Temporal Video Grounding dataset by utilizing these datasets. Figure \ref{fig:dataset1} shows the pipeline of data synthesis, which contains four components: i) Analyzer for object extraction. ii) Annotator for box generation. iii) Refiner for time boundary. iv) Filter for bounding box.

\noindent\textbf{Analyzer for Object Extraction.} We use Qwen2.5-72B~\cite{ref:qwen2.5} as an analyzer to extract objects from the captions of each video, identifying objects in the captions that can be localized, such as people, animals, vehicles, etc. In subsequent processes, we prioritize locating the subject of the caption, followed by other mentioned objects. 

\noindent\textbf{Annotator for Box Generation.} We annotate the bounding box for each frame within the timestamps range to ensure precise spatial localization. Grounding-DINO~\cite{ref:groundingdino} is leveraged to extract the bounding box for open-set grounding, with the extracted object as the text prompt. It allows us to generate multiple high-confidence bounding boxes, filtering out those that fall below the 0.3 threshold.

\noindent\textbf{Refiner for Time Boundary.} The timestamp annotations in video datasets are not always precise. For instance, DiDeMo~\cite{ref:didemo} adopts a time interval that is an integer multiple of five, leading to many start and end times that do not correspond to any actual objects. To address this issue, we refine the temporal boundaries by adjusting timestamps to better align with actual object appearances based on the Grounding-DINO's output. Additionally, we filter out adjusted timestamps that are either shorter than 2 seconds or longer than 120 seconds.

\noindent\textbf{Filter for Bounding Box.} Obtaining high-quality bounding boxes for each frame is crucial for precise spatial-temporal localization. To refine our annotations, we implement a multi-step filtering process. First, filter the complex scenes. If a frame contains more than three bounding boxes, we classify it as a complex scene and discard it. For the remaining frames, we retain only the bounding box with the highest confidence score. Second, filter through the object size. Object sizes should not fluctuate drastically between consecutive frames. To enforce this, we compare the bounding box area of each frame with that of the bounding box across adjacent frames. We remove samples where the box area is less than half or more than twice the reference box’s area, ensuring stable object localization. Through this filtering pipeline, we eliminate approximately 40\% of samples, resulting in a high-quality spatial-temporal dataset for fine-grained video understanding.

\begin{table*}[ht]
	\centering
	\begin{minipage}{.5\textwidth}
		\centering
		\renewcommand{\arraystretch}{1.0}
		\scalebox{0.80}{
			\begin{tabular}{lcccc}
				\specialrule{1.5pt}{0pt}{0pt}
				Models & m\_tIoU & m\_vIoU & vIoU@0.3 &  vIoU@0.5  \\ 
				\midrule 
                \multicolumn{5}{c}{\textit{Non-generative and task-specific models}} \\
				STGVT~\cite{ref:hcstvg} & - &  18.2 & 26.8 & 9.5  \\
				STVGBert~\cite{ref:stvgbert} & - & 20.4 & 29.4 &  11.3  \\
				TubeDETR~\cite{ref:tubedetr} & 43.7 & 32.4 & 49.8 & 23.5 \\
				STCAT~\cite{ref:embrace} & 49.4 & 35.1 & 57.7 & 30.1 \\
				STVGFormer~\cite{ref:coll} & - & 36.9 & \textbf{62.2} & 34.8 \\ 
				
				CG-STVG~\cite{ref:cgstvg} & \textbf{52.8} & \textbf{38.4} & 61.5 & \textbf{36.3} \\ \midrule 
                \multicolumn{5}{c}{\textit{Video LLMs with Parameter Sizes of 7B}}\\
                Qwen2.5-VL-7B~\cite{ref:qwen2.5} & 25.6 &  19.1& 20.2 &12.6  \\
                GroundingGPT-7B~\cite{ref:groundinggpt}&22.2  &16.7  &15.0 &4.9  \\
                TRACE-7B~\cite{ref:trace}&39.2  &-  &-&-  \\
                \midrule 
                \textit{SpaceVLLM}-7B & \textbf{56.9} & \textbf{39.3} & \textbf{66.6} & \textbf{36.9} \\ 
                \specialrule{1.5pt}{0pt}{0pt}
		\end{tabular}}
		\caption{Comparison with others on HCSTVG-v1 test set  (\%).}
		\label{tab:hcstvgv1}
	\end{minipage}%
	\hfill
	\begin{minipage}{.5\textwidth}
		\centering
		\renewcommand{\arraystretch}{1.0}
		\scalebox{0.80}{
			\begin{tabular}{lcccc}
				\specialrule{1.5pt}{0pt}{0pt}
				Models & m\_tIoU & m\_vIoU & vIoU@0.3 &  vIoU@0.5  \\ \midrule 
                \multicolumn{5}{c}{\textit{Non-generative and task-specific models}} \\
				PCC~\cite{yu20212rd} & - &  30.0 & - & -  \\ 
				2D-Tan~\cite{tan2021augmented}  & - & 30.4 &  50.4 & 18.8  \\
				MMN~\cite{wang2022negative} & - & 30.3 & 49.0 & 25.6 \\
				TubeDETR~\cite{ref:tubedetr} & - & 36.4 & 58.8 & 30.6 \\
				STVGFormer~\cite{ref:coll} & 58.1 & 38.7 & \textbf{65.5} & 33.8 \\ 
				
				CG-STVG~\cite{ref:cgstvg} & \textbf{60.0} & \textbf{39.5} & 64.5 & \textbf{36.3} \\ \midrule
                \multicolumn{5}{c}{\textit{Video LLMs with Parameter Sizes of 7B}}\\
                Qwen2.5-VL-7B~\cite{ref:qwen2.5} &22.9  &13.0  &15.6  &6.4  \\
                GroundingGPT-7B~\cite{ref:groundinggpt}&19.6  &14.7  &16.6 &3.1 \\
                TRACE-7B~\cite{ref:trace}&43.8  &-  &-&-  \\
                \midrule 
                \textit{SpaceVLLM}-7B & \textbf{58.0} & \textbf{34.0} & \textbf{56.9} & \textbf{24.7} \\ 
				\specialrule{1.5pt}{0pt}{0pt}
		\end{tabular}}
		\caption{Comparison with others on HCSTVG-v2 val. set  (\%).}
		\label{tab:hcstvgv2}
	\end{minipage}
\end{table*}
\begin{table*}[htb]
	\centering
	\renewcommand{\arraystretch}{1.0}
	\scalebox{0.85}{
		\begin{tabular}{lcccccccc}
			\specialrule{1.5pt}{0pt}{0pt}
                 & \multicolumn{4}{c}{  Declarative Sentences} & \multicolumn{4}{c}{ Interrogative Sentences} \\ 
			\multirow{-2}{*}{ Models} & m\_tIoU & m\_vIoU & vIoU@0.3 &  vIoU@0.5  & m\_tIoU & m\_vIoU & vIoU@0.3 &  vIoU@0.5  \\
			\midrule
            \multicolumn{9}{c}{\textit{Non-generative and task-specific models}} \\
			STGRN~\cite{ref:vidstvg}  &  48.5 &  19.8 & 25.8 & 14.6 &  47.0 & 18.3 & 21.1 & 12.8 \\
			OMRN~\cite{ref:object}  &  50.7 &  23.1 & 32.6 & 16.4 &  49.2 & 20.6 & 28.4 & 14.1 \\
			STGVT~\cite{ref:hcstvg} & - &  21.6 & 29.8 & 18.9 &  - & - & -  & - \\
			STVGBert~\cite{ref:stvgbert}  & - &  24.0 & 30.9 & 18.4 & - & 22.5 & 26.0 & 16.0 \\
			TubeDETR~\cite{ref:tubedetr} & 48.1 &  30.4 & 42.5 & 28.2 & 46.9 & 25.7 & 35.7 & 23.2 \\
			STCAT~\cite{ref:embrace} & 50.8 & 33.1 & 46.2 & 32.6 & 49.7 & 28.2 & 39.2 & 26.6  \\
			STVGFormer~\cite{ref:coll} & - & 33.7 & 47.2 & 32.8 & - & 28.5 & 39.9 & 26.2  \\ 
            CG-STVG~\cite{ref:cgstvg} & \textbf{51.4} & \textbf{34.0} & \textbf{47.7} & \textbf{33.1} &  \textbf{49.9} & \textbf{29.0} & \textbf{40.5} & \textbf{27.5} \\ \midrule
            \multicolumn{9}{c}{\textit{Video LLMs with Parameter Sizes of 7B}} \\
             Qwen2.5-VL-7B~\cite{ref:qwen2.5}& {16.8} & {10.9} & {14.3} & {5.4} &  {13.8} & {8.5} & {11.3} & {4.4} \\
            GroundingGPT-7B~\cite{ref:groundinggpt} & {15.5} & {12.3} & {13.2} & {4.1} &  {11.9} & {8.7} & {9.6} & {2.9}\\
            TRACE-7B~\cite{ref:trace} & {24.3} & - & - & - &  20.2 & - & - & -\\
            \midrule
			\textit{SpaceVLLM}-7B  & \textbf{47.7} & \textbf{27.4} & \textbf{39.1} & \textbf{26.2} & \textbf{48.5} & \textbf{25.4} & \textbf{35.9} & \textbf{22.2}  \\
			\specialrule{1.5pt}{0pt}{0pt}
	\end{tabular}}
	\caption{Comparison with existing state-of-the-art models on VidSTG test set (\%).}
	\label{tab:vidstg}
	\vspace{-2mm}
\end{table*}

\subsubsection{Model Training}
\label{sec:strategy}
To equip the model with spatio-temporal localization capabilities while preserving its general understanding abilities, we merge different tasks for multi-task instruction tuning. For the multi-task instruction tuning, we apply the Uni-STG dataset to endow the MLLMs with the spatio-temporal capability. Additionally, to maintain the model’s general understanding abilities, we merge Uni-STG dataset along with Visual Question Answering, Conversation and Video Captioning, which ensures the model not only excels in fine-grained spatial-temporal localization but also maintains general video understanding. Table \ref{tab:dataset} shows the training datasets and tasks to enhance fine-grained understanding.

\section{Experiments}
\label{sec:exp}

\subsection{Experimental Setting}
\paragraph{Implementation Details.} 
We employed SigLIP~\cite{ref:siglip} as our vision encoder, and Qwen2~\cite{yang2024qwen2} as the LLM. We use AdamW~\citep{ref:loshchilov2017decoupled} optimizer with the learning rate and weight decay set to 1e-5 and 0, respectively. We also adopt cosine as the learning rate scheduler, where the warmup ratio is set to 0.03. We train the \textit{SpaceVLLM} with 16 NVIDIA A800 GPUs in 24 hours based on the LLaVA-Video~\cite{ref:llavavideo} model. 
The weights of the time loss $\lambda_{time}$ and the space loss $\lambda_{space}$ are set to $1.0$ and $1.0$, respectively, and those of the $\mathcal{L}_1$ loss $\lambda_{\mathcal{L}_1}$ and the generalized IoU loss $\lambda_{giou}$ are set to $3.0$ and $1.0$, respectively.

\noindent \textbf{Evaluation Benchmarks.}
For a comprehensive evaluation, we consider 11 benchmarks that cover Spatio-Temporal Video Grounding (STVG), Video Temporal Grounding (VTG), Referring Expression Comprehension (REC), and various video understanding tasks. Following~\cite{ref:stvgbert, ref:tubedetr, ref:embrace}, we conduct spatio-temporal video grounding experiments on three benchmarks, including HCSTVG-v1~\cite{ref:hcstvg}, HCSTVG-v2~\cite{ref:hcstvg}, and VidSTG~\cite{ref:vidstvg}. For referring expression comprehension, we use the RefCOCO~\cite{ref:refcoco}, RefCOCO+~\cite{ref:refcoco}, and RefCOCOg~\cite{ref:refcocog} datasets. We also use Charades-STA~\cite{ref:charades} for video temporal grounding. Additional evaluations include the MVBench~\cite{ref:mvbench}, VideoMME~\cite{ref:videomme}, TempCompass~\cite{ref:tempcompass}, and EgoSchema~\cite{ref:egoschema} for video understanding. To fairly compare with other models, we primarily use results from original papers. When such results are not available, we assess the models using LMMs-Eval~\cite{zhang2024lmms} or official scripts.

\noindent \textbf{Evaluation Metrics of STVG.}
Following~\cite{ref:stvgbert,ref:tubedetr, ref:embrace}, we use m\_tIoU, m\_vIoU, and vIoU@R as evaluation metrics for STVG. m\_tIoU measures temporal localization performance, while m\_vIoU and vIoU@R evaluate spatial localization. 

\subsection{Performance of \textit{\textbf{SpaceVLLM}}}

\subsubsection{Spatio-Temporal Video Grounding Task}
To demonstrate the effectiveness of \textit{SpaceVLLM}, we compare it to state-of-the-art task-specific models and MLLMs capable of temporal or image grounding. It is worth noting that among the MLLMs used for comparison, TRACE~\cite{ref:trace} focuses on metrics related to temporal grounding, while Qwen2.5-VL~\cite{ref:qwen2.5} directly generates timestamps and spatial coordinate by devising prompts and GroundingGPT~\cite{ref:groundinggpt} uses a two-stage approach for evaluation. Specifically, it first performs temporal grounding based on the text caption to determine the frame range and then carries out image spatial grounding on each frame within that range.

\noindent \textbf{HCSTVG-v1 and HCSTVG-v2.} Table~\ref{tab:hcstvgv1} shows the results on the HCSTVG-v1 test set, and our proposed method achieves state-of-the-art performance in all metrics. 
When compared to other video LLMs, \textit{SpaceVLLM} attains 39.3 m\_vIoU scores outperforming Qwen2.5-VL~\cite{ref:qwen2.5} by 20.2, and achieves 56.9 m\_tIoU scores surpassing TRACE~\cite{ref:trace} by 17.7. Significantly, in comparison with CG-STVG~\cite{ref:cgstvg} which is the state-of-the-art DETR-like architecture model, our method improves the 4.1, 0.9, 5.1, and 0.6 absolute scores on m\_tIoU, m\_vIoU, vIoU@0.3 and vIoU@0.5 metrics, respectively. On the more comprehensive validation set of the HCSTVG-v2, our method also performs excellently in the four metrics as illustrated in Table~\ref{tab:hcstvgv2}. Specifically, our method improves the 19.3 absolute m\_vIoU score compared to GroundingGPT~\cite{ref:groundinggpt} and improves 14.2 absolute m\_tIoU score compared to TRACE~\cite{ref:trace}.
Furthermore, \textit{SpaceVLLM}'s performance on the HCSTVG-v2 is also competitive with traditional non-generative and task-specific methods such as CG-STVG~\cite{ref:cgstvg}, STVGFormer~\cite{ref:coll}, and TubeDETR~\cite{ref:tubedetr}. However, these methods cannot handle multiple tasks simultaneously.

\noindent \textbf{VidSTG.} We evaluate the performance of \textit{SpaceVLLM} on the more challenging VidSTG datasets in Table~\ref{tab:vidstg}. Unlike HCSTVG's declarative-only annotation, the text captions in VidSTG include both declarative and interrogative sentences. As shown, Qwen2.5-VL~\cite{ref:qwen2.5} and GroundingGPT~\cite{ref:groundinggpt} perform poorly in the declarative sentences section, scoring 10.9 and 12.3 m\_vIoU scores, respectively. In the interrogative sentences section, where extracting targets based on video information is required, they score only 8.5 and 8.7 in m\_vIoU. In contrast, our method significantly outperforms TRACE~\cite{ref:trace} by 23.4 in m\_tIoU for declarative sentences and 28.3 for interrogative sentences. Moreover, our model maintains performance comparable to task-specific models, demonstrating its robustness in extracting accurate and effective spatio-temporal information even when faced with more complex scenarios.

\subsubsection{Video Temporal Grounding Task}
Table \ref{tab:charades_sta} presents the performance on the Charades-STA\cite{ref:charades} dataset for the task of Video Temporal Grounding. We compare our model with several traditional state-of-the-art models and LLM-based models that have been fine-tuned using this dataset. Our method achieves state-of-the-art performance in R@1$_{\text{IoU}=0.5}$ among these models. When compared to the current best LLM-based model, TRACE \cite{ref:trace}, which focuses on video temporal grounding, our model still demonstrates comparable results and even surpasses it by 1.9\% in R@1$_{\text{IoU}=0.5}$. This is because our method facilitates the MLLM in extracting valuable temporal and spatial information during training, thereby enhancing the \textit{SpaceVLLM}'s video temporal grounding capability.

\subsubsection{Referring Expression Comprehension Task}
To independently validate spatial capability, we conduct experiments on the Referring Expression Comprehension task. 
As depicted in Table~\ref{tab:refcoco}, our model achieves state-of-the-art performance on 8 metrics. Compared to the previous best model~\cite{ref:groma}, \textit{SpaceVLLM} improves by 2.4\% on the RefCOCO+~\cite{ref:refcoco} validation set, by 2\% on the test-A set, and by 1.7\% on the test-B set. Since the image spatial-aware query only needs to learn spatial information, these outstanding results demonstrate that our method can accurately preserve beneficial spatial details for localization.
\begin{table}[t]
\centering
\renewcommand{\arraystretch}{1.055}
\scalebox{1}{
\begin{tabular}{lcc}
\specialrule{1.5pt}{0pt}{0pt}
\multirow{2}{*}{\text{Models}} & \multicolumn{2}{c}{{Charades-STA}} \\ 
\cmidrule(lr){2-3} 
 & R@1$_{\text{IoU}=0.5}$ & R@1$_{\text{IoU}=0.7}$ \\
\midrule
\multicolumn{3}{c}{\textit{Traditional models}} \\
Moment-DETR~\cite{ref:momentdetr} & 55.7 & 34.2 \\
QD-DETR~\cite{ref:qddetr} & 57.3 & 32.6 \\
MomentDiff~\cite{ref:momentdiff} & 55.6 & 32.4 \\
\midrule
\multicolumn{3}{c}{\textit{Video LLMs with Parameter Sizes of 7B}} \\
HawkEye-7B~\cite{ref:hawkeye} & 58.3 & 28.8 \\
TimeChat-7B~\cite{ref:timechat} & 46.7 & 23.7 \\
VTG-LLM-7B~\cite{ref:vtgllm} & 57.2 & 33.4 \\
TRACE-7B~\cite{ref:trace} & 61.7 & \textbf{41.4} \\
\midrule
\textit{SpaceVLLM}-7B & \textbf{63.6} & 38.5 \\
\specialrule{1.5pt}{0pt}{0pt}
\end{tabular}}
\caption{Results on Charades-STA for Video Temporal Grounding.}
\label{tab:charades_sta}
\vspace{-4mm}
\end{table}

\subsubsection{Video Understanding Task}
In addition to \textit{SpaceVLLM}'s strong spatio-temporal capabilities, we also evaluate our model on four video understanding benchmarks: MVBench \cite{ref:mvbench}, VideoMME \cite{ref:videomme}, EgoSchema \cite{ref:egoschema}, and TempCompass \cite{ref:tempcompass}, as shown in Table~\ref{tab:VQA_new}. Compared to the base model LLaVA-Video~\cite{ref:llavavideo}, our model achieved improvements of 0.7\%, 0.1\%, and 0.3\% on MVBench, EgoSchema, and TempCompass, respectively. These results validate that our model still maintains a strong ability for video understanding.

\subsection{Ablation Study}
In this section, we present ablation studies on \textit{SpaceVLLM}. Specifically, we evaluate the effectiveness of our proposed module in Table \ref{tab:ab_modules} and analyze the impact of the number of queries in Table \ref{tab:ab_queries}. All experiments are conducted on the HCSTVG-v1~\cite{ref:hcstvg} to assess the model's performance.

\noindent\textbf{Model Architecture.} As presented in Table~\ref{tab:ab_modules}, the first row represents the model without the two modules, where it is trained to directly output timestamps along with all coordinates. This results in a significant 11.6\% performance drop on the metric of m\_vIoU and 3.1\% decline on the metric of m\_tIoU, highlighting the challenge of temporal-spatial misalignment when MLLMs attempt to simultaneously capture both dimensions. 
\begin{table}[t]
    \centering
    \renewcommand{\arraystretch}{1.17}
    \resizebox{0.5\textwidth}{!}{
    \begin{tabular}{l|ccc|ccc|ccc}
        \specialrule{1.5pt}{0pt}{0pt}
        \multirow{2}{*}{Models} & \multicolumn{3}{c}{RefCOCO} & \multicolumn{3}{c}{RefCOCO+} & \multicolumn{2}{c}{RefCOCOg} \\
        \cmidrule(lr){2-4} \cmidrule(lr){5-7} \cmidrule(lr){8-9}
        & val & test-A & test-B & val & test-A & test-B & val-u & test-u \\
        \midrule
        Shikra-7B~\cite{ref:shikra} & 87.0 & 90.6 & 80.2 & 81.6 & 87.4 & 72.1 & 82.3 & 82.2 \\
        Ferret-7B~\cite{ref:fer} & 87.5 & 91.4 & 82.5 & 80.8 & 87.4 & 73.1 & 83.9 & 84.8 \\
        GroundingGPT-7B~\cite{ref:groundinggpt} & 88.0 & 91.6 & 82.5 & 81.6 & 87.2 & 73.2 & 81.7 & 82.0 \\
        MiniGPT-v2-7B~\cite{ref:mini} & 88.7 & 91.7 & 85.3 & 80.0 & 85.1 & 74.5 & 84.4 & 84.7 \\
        Elysium-7B~\cite{ref:elysium} & 89.1 & 92.1 & 85.0 & 82.9 & 88.9 & 75.6 & 82.9 & 83.6 \\
        Groma-7B~\cite{ref:groma} & 89.5 & 92.1 & 86.3 & 83.9 & 88.9 & 78.1 & 86.3 & 87.0 \\
        \midrule
        \textit{SpaceVLLM}-7B & \textbf{90.8} & \textbf{93.4} & \textbf{87.0} & \textbf{86.3} & \textbf{90.9} & \textbf{79.8} & \textbf{86.8} & \textbf{88.0} \\
        \specialrule{1.5pt}{0pt}{0pt}
    \end{tabular}  }
    \caption{Performance comparison of various models on RefCOCO, RefCOCO+, and RefCOCOg benchmarks (\%). The accuracy with IoU threshold is 0.5.}
    \label{tab:refcoco}
\end{table}

\begin{table}[t]
\centering
\renewcommand{\arraystretch}{1.1}
\resizebox{0.5\textwidth}{!}{
\begin{tabular}{lcccc}
\specialrule{1.5pt}{0pt}{0pt}
Methods & m\_tIoU & m\_vIoU & vIoU@0.3 & vIoU@0.5 \\ 
\midrule
w/o Queries and Decoder & 53.8 & 27.7 & 45.5 & 22.4 \\
w/o Interleaved Design & 55.5 & 32.9 & 52.5 & 28.4 \\
w/o Dual Cross Attention & 56.5 & 36.7 & 59.6 & 33.3 \\
\midrule
\textit{SpaceVLLM} & \textbf{56.9} & \textbf{39.3} & \textbf{66.6} & \textbf{36.9} \\ 
\specialrule{1.5pt}{0pt}{0pt}
\end{tabular} }
\caption{Ablation studies on the modules of \textit{SpaceVLLM}, with evaluation on HCSTVG-v1.}
\label{tab:ab_modules}
\vspace{-3mm}
\end{table}
The second row shows the performance when all the queries are concatenated after the visual tokens, instead of being interleaved between the visual tokens of each frame. The 6.4\% performance degradation demonstrates the effectiveness of our interleaved design in capturing dynamic spatial information. The third row reveals that removing the dual cross-attention module leads to a significant decline from 39.3\% to 36.7\%, emphasizing the importance of cross-attention in strengthening the spatial connections between queries and visual representations. 

\noindent\textbf{Number of queries.} Table \ref{tab:ab_queries} reports the results of spatio-temporal localization under different numbers of queries. A smaller number of queries fails to capture sufficient temporal and spatial information across frames, resulting in suboptimal performance in both aspects. Conversely, increasing the number of queries slightly improves localization but leads to significantly higher memory cost. We finally set the number of queries to 64 as a balanced trade-off between performance and computational efficiency.
\definecolor{customGreen}{RGB}{33, 221, 33} 
\definecolor{customPurple}{RGB}{87, 53, 227} 
\definecolor{customYellow}{RGB}{255, 212, 90} 
\definecolor{customRed}{RGB}{255,0,0} 
\begin{table}[t]
\centering
\resizebox{0.5\textwidth}{!}{
\begin{tabular}{lccccc}
\specialrule{1.5pt}{0pt}{0pt}
{\multirow{2}{*}{Models}} & \multicolumn{1}{c}{\multirow{2}{*}{MVBench}} & \multicolumn{1}{c}{\multirow{2}{*}{EgoSchema}} & \multicolumn{1}{c}{\multirow{2}{*}{TempCompass}} & \multicolumn{2}{c}{VideoMME} \\
       &   &   &   & \multicolumn{2}{c}{(wo/w-subs)}  \\
\midrule
Timechat-7B~\cite{ref:timechat} & 38.5 & 33.0 & 50.7 & 34.7/ - \\
Videochat2-7B~\cite{ref:mvbench} & 51.1 & 54.4 & 51.1 & 42.3 / 54.6 \\
Video-LLaVA-7B~\cite{ref:videollava} & 43.0 & 40.7 & 45.6 & 39.9 / 41.6 \\
LongVA-7B~\cite{ref:longva} & - & 44.1 & 56.1 & 52.6 / 54.3 \\
Videollama2.1-7B~\cite{ref:videollama2} & 57.3 & 53.1 & 56.8 & 54.9 / 56.4 \\
LLaVA-OV-7B~\cite{ref:llavaov} & 56.7 &\textbf{60.5} & 63.6 & 58.2 / 61.5 \\
LLaVA-Video-7B~\cite{ref:llavavideo} & 58.6 & 57.3 & 67.0 & \textbf{63.3} / \textbf{69.7} \\
\midrule
\textit{SpaceVLLM}-7B  & \textbf{59.3} & {57.4} & \textbf{67.3} & 60.0 / 65.6 \\
\specialrule{1.5pt}{0pt}{0pt}
\end{tabular} }
\caption{Performance comparison of various models on general video understanding tasks, evaluated across benchmarks such as MVBench, EgoSchema, TempCompass, and VideoMME (\%).}
\label{tab:VQA_new}
\vspace{-1mm}
\end{table}

\begin{table}[t]
\centering
\resizebox{0.5\textwidth}{!}{
\begin{tabular}{ccccc}
\specialrule{1.5pt}{0pt}{0pt}
Number of queries & m\_tIoU & m\_vIoU & vIoU@0.3 & vIoU@0.5 \\ 
\midrule
32  & 55.2 & 37.0 & 63.8 & 35.0 \\
64  & 56.9 & 39.3 & 66.6 & 36.9 \\
96  & \textbf{58.0} & \textbf{40.2} & \textbf{67.8} & \textbf{37.6} \\
\specialrule{1.5pt}{0pt}{0pt}
\end{tabular} }
\caption{Ablation studies on the numbers of queries, evaluated on HCSTVG-v1.}
\label{tab:ab_queries}
\end{table}
\begin{figure}[t]
\centering
\includegraphics[width=\linewidth]{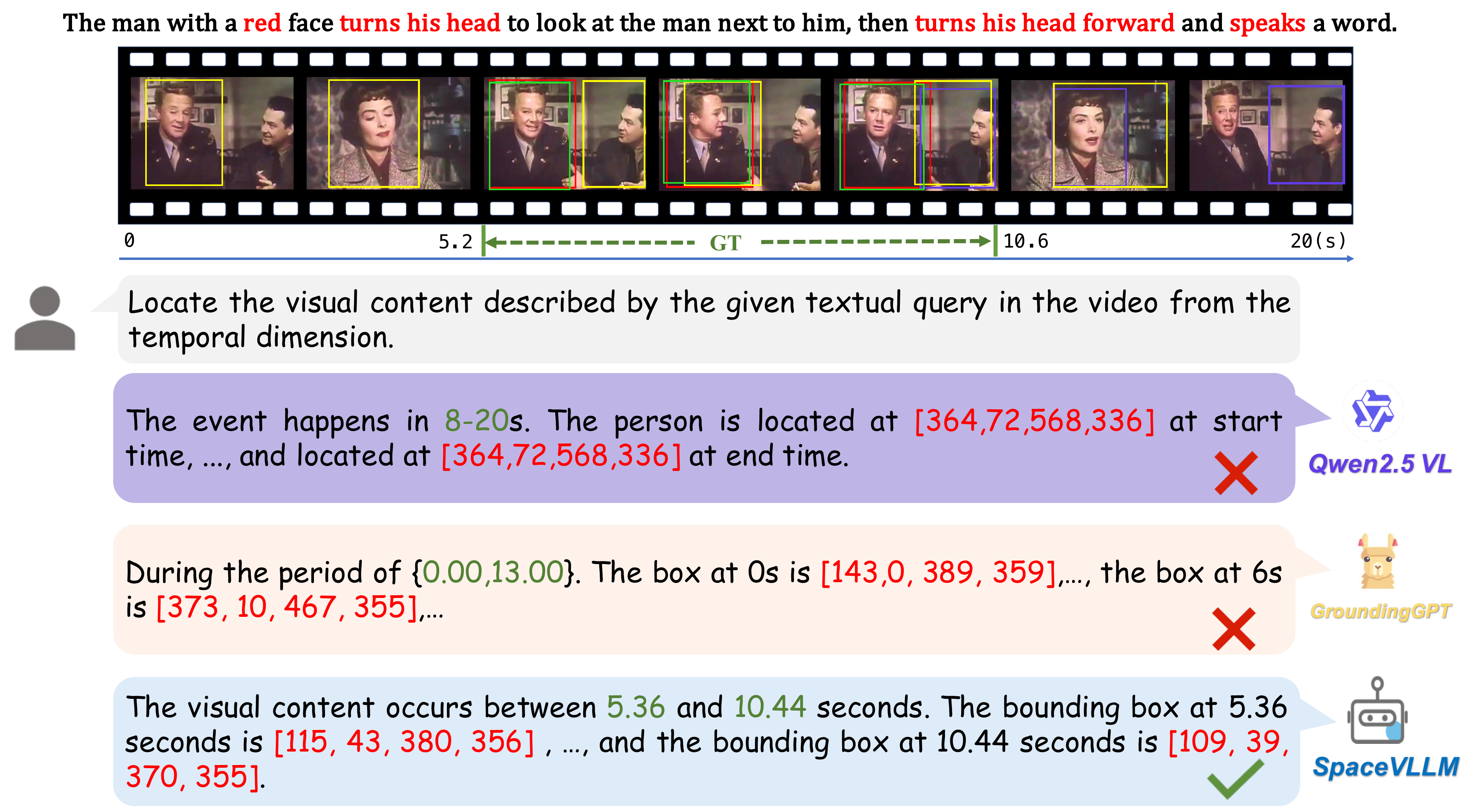}
\caption{Visualization between LLM-based model for the task of Spatio-Temporal Video Grounding. As for the box in the video, \textcolor{customGreen}{green} is the ground-truth bounding box, \textcolor{customPurple}{purple} is the Qwen 2.5 VL, \textcolor{customYellow}{yellow} is the GroundingGPT and \textcolor{customRed}{red} is our \textit{SpaceVLLM}.}
\label{fig:visual1}
\vspace{-10pt}
\end{figure}

\subsection{Visualization}  
In this section, we compare our model with two other models, Qwen2.5-VL~\cite{ref:qwen2.5} and GroundingGPT~\cite{ ref:groundinggpt} for Spatio-Temporal Video Grounding. These models are limited to capturing accurate spatio-temporal details and aligning visual tokens of each frame with their corresponding coordinates. As shown in Figure \ref{fig:visual1}, it is challenging to distinguish which person is being referred to given the text query, such as ``turn his head" or ``turn his head forward" for a single frame. In contrast, our model can accurately localize the object, as our spatio-temporal aware queries retain abundant temporal perception and dynamic spatial information, which is crucial for fine-grained understanding. 
\section{Conclusion}
\label{sec:conclusion}

In this paper, we introduce \textit{SpaceVLLM}, a MLLM with spatio-temporal video grounding capability. We first insert a set of interleaved spatio-temporal aware queries after each sampled frame of visual tokens. Secondly, we develop a Query-Guided Space Decoder that effectively links queries with spatio-temporal information to spatial coordinates. Moreover, we introduce a Unified Spatio-Temporal Grounding (Uni-STG) dataset to advance multimodal spatio-temporal understanding. Extensive experiments demonstrate that our model achieves state-of-the-art performance across 11 benchmarks, including temporal, spatial, spatio-temporal, and multimodal understanding tasks, fully validating the effectiveness of our model.

{
    \small
    \bibliographystyle{ieeenat_fullname}
    \bibliography{main}

\begin{thebibliography}{61}
\providecommand{\natexlab}[1]{#1}
\providecommand{\url}[1]{\texttt{#1}}
\expandafter\ifx\csname urlstyle\endcsname\relax
  \providecommand{\doi}[1]{doi: #1}\else
  \providecommand{\doi}{doi: \begingroup \urlstyle{rm}\Url}\fi

\bibitem[Chen et~al.(2023{\natexlab{a}})Chen, Zhu, Shen, Li, Liu, Zhang, Krishnamoorthi, Chandra, Xiong, and Elhoseiny]{ref:mini}
Jun Chen, Deyao Zhu, Xiaoqian Shen, Xiang Li, Zechun Liu, Pengchuan Zhang, Raghuraman Krishnamoorthi, Vikas Chandra, Yunyang Xiong, and Mohamed Elhoseiny.
\newblock Minigpt-v2: large language model as a unified interface for vision-language multi-task learning.
\newblock \emph{arXiv preprint arXiv:2310.09478}, 2023{\natexlab{a}}.

\bibitem[Chen et~al.(2023{\natexlab{b}})Chen, Zhang, Zeng, Zhang, Zhu, and Zhao]{ref:shikra}
Keqin Chen, Zhao Zhang, Weili Zeng, Richong Zhang, Feng Zhu, and Rui Zhao.
\newblock Shikra: Unleashing multimodal llm's referential dialogue magic.
\newblock \emph{arXiv preprint arXiv:2306.15195}, 2023{\natexlab{b}}.

\bibitem[Chen et~al.(2024)Chen, Wei, Li, Dong, Zhang, Zang, Chen, Duan, Lin, Tang, et~al.]{ref:chen2024sharegpt4video}
Lin Chen, Xilin Wei, Jinsong Li, Xiaoyi Dong, Pan Zhang, Yuhang Zang, Zehui Chen, Haodong Duan, Bin Lin, Zhenyu Tang, et~al.
\newblock Sharegpt4video: Improving video understanding and generation with better captions.
\newblock \emph{arXiv preprint arXiv:2406.04325}, 2024.

\bibitem[Cheng et~al.(2024)Cheng, Leng, Zhang, Xin, Li, Chen, Zhu, Zhang, Luo, Zhao, et~al.]{ref:videollama2}
Zesen Cheng, Sicong Leng, Hang Zhang, Yifei Xin, Xin Li, Guanzheng Chen, Yongxin Zhu, Wenqi Zhang, Ziyang Luo, Deli Zhao, et~al.
\newblock Videollama 2: Advancing spatial-temporal modeling and audio understanding in video-llms.
\newblock \emph{arXiv preprint arXiv:2406.07476}, 2024.

\bibitem[Fu et~al.(2024)Fu, Dai, Luo, Li, Ren, Zhang, Wang, Zhou, Shen, Zhang, et~al.]{ref:videomme}
Chaoyou Fu, Yuhan Dai, Yongdong Luo, Lei Li, Shuhuai Ren, Renrui Zhang, Zihan Wang, Chenyu Zhou, Yunhang Shen, Mengdan Zhang, et~al.
\newblock Video-mme: The first-ever comprehensive evaluation benchmark of multi-modal llms in video analysis.
\newblock \emph{arXiv preprint arXiv:2405.21075}, 2024.

\bibitem[Gao et~al.(2017)Gao, Sun, Yang, and Nevatia]{ref:charades}
Jiyang Gao, Chen Sun, Zhenheng Yang, and Ram Nevatia.
\newblock Tall: Temporal activity localization via language query.
\newblock In \emph{Proceedings of the IEEE International Conference on Computer Vision}, pages 5267--5275, 2017.

\bibitem[Girshick(2015)]{ref:fasterrcnn}
Ross Girshick.
\newblock Fast r-cnn.
\newblock In \emph{Proceedings of the IEEE International Conference on Computer Vision}, pages 1440--1448, 2015.

\bibitem[Gu et~al.(2024)Gu, Fan, Huang, Luo, and Zhang]{ref:cgstvg}
Xin Gu, Heng Fan, Yan Huang, Tiejian Luo, and Libo Zhang.
\newblock Context-guided spatio-temporal video grounding.
\newblock In \emph{Proceedings of the IEEE/CVF Conference on Computer Vision and Pattern Recognition}, pages 18330--18339, 2024.

\bibitem[Guo et~al.(2024{\natexlab{a}})Guo, Liu, Li, Cheng, Tang, Sui, Liu, Chen, and Zhao]{ref:vtgllm}
Yongxin Guo, Jingyu Liu, Mingda Li, Dingxin Cheng, Xiaoying Tang, Dianbo Sui, Qingbin Liu, Xi Chen, and Kevin Zhao.
\newblock Vtg-llm:integrating timestamp knowledge into video llms for enhanced video temporal grounding.
\newblock \emph{arXiv preprint arXiv:2405.13382}, 2024{\natexlab{a}}.

\bibitem[Guo et~al.(2024{\natexlab{b}})Guo, Liu, Li, Tang, Liu, and Chen]{ref:trace}
Yongxin Guo, Jingyu Liu, Mingda Li, Xiaoying Tang, Qingbin Liu, and Xi Chen.
\newblock Trace: Temporal grounding video llm via causal event modeling.
\newblock \emph{arXiv preprint arXiv:2410.05643}, 2024{\natexlab{b}}.

\bibitem[Hendricks et~al.(2017)Hendricks, Wang, Shechtman, Sivic, Darrell, and Russell]{ref:didemo}
Lisa~Anne Hendricks, Oliver Wang, Eli Shechtman, Josef Sivic, Trevor Darrell, and Bryan Russell.
\newblock Localizing moments in video with natural language.
\newblock In \emph{Proceedings of the IEEE International Conference on Computer Vision}, pages 5803--5812, 2017.

\bibitem[Huang et~al.(2024)Huang, Wang, Chen, Song, and Zhu]{ref:vtimellm}
Bin Huang, Xin Wang, Hong Chen, Zihan Song, and Wenwu Zhu.
\newblock Vtimellm: Empower llm to grasp video moments.
\newblock In \emph{Proceedings of the IEEE/CVF Conference on Computer Vision and Pattern Recognition}, pages 14271--14280, 2024.

\bibitem[Jin et~al.(2022)Jin, yongzhi li, Yuan, and Mu]{ref:embrace}
Yang Jin, yongzhi li, Zehuan Yuan, and Yadong Mu.
\newblock Embracing consistency: A one-stage approach for spatio-temporal video grounding.
\newblock In \emph{Advances in Neural Information Processing Systems}, pages 29192--29204, 2022.

\bibitem[Kazemzadeh et~al.(2014)Kazemzadeh, Ordonez, Matten, and Berg]{ref:refcoco}
Sahar Kazemzadeh, Vicente Ordonez, Mark Matten, and Tamara~L. Berg.
\newblock Referitgame: Referring to objects in photographs of natural scenes.
\newblock In \emph{Proceedings of the 2014 conference on empirical methods in natural language processing}, page 787–798, 2014.

\bibitem[Krishna et~al.(2016)Krishna, Zhu, Groth, Johnson, Hata, Kravitz, Chen, Kalantidis, Li, Shamma, Bernstein, and Fei-Fei]{ref:visual}
Ranjay Krishna, Yuke Zhu, Oliver Groth, Justin Johnson, Kenji Hata, Joshua Kravitz, Stephanie Chen, Yannis Kalantidis, Li-Jia Li, David~A. Shamma, Michael~S. Bernstein, and Li Fei-Fei.
\newblock Visual genome: Connecting language and vision using crowdsourced dense image annotations.
\newblock \emph{International journal of computer vision}, 123, 2016.

\bibitem[Lei et~al.(2021)Lei, Berg, and Bansal]{ref:momentdetr}
Jie Lei, Tamara~L. Berg, and Mohit Bansal.
\newblock Detecting moments and highlights in videos via natural language queries.
\newblock In \emph{Advances in neural information processing systems}, pages 11846--11858, 2021.

\bibitem[Li et~al.(2024{\natexlab{a}})Li, Zhang, Guo, Zhang, Li, Zhang, Zhang, Zhang, Li, Liu, et~al.]{ref:llavaov}
Bo Li, Yuanhan Zhang, Dong Guo, Renrui Zhang, Feng Li, Hao Zhang, Kaichen Zhang, Peiyuan Zhang, Yanwei Li, Ziwei Liu, et~al.
\newblock Llava-onevision: Easy visual task transfer.
\newblock \emph{arXiv preprint arXiv:2408.03326}, 2024{\natexlab{a}}.

\bibitem[Li et~al.(2023{\natexlab{a}})Li, He, Wang, Li, Wang, Luo, Wang, Wang, and Qiao]{ref:videochat}
KunChang Li, Yinan He, Yi Wang, Yizhuo Li, Wenhai Wang, Ping Luo, Yali Wang, Limin Wang, and Yu Qiao.
\newblock Videochat: Chat-centric video understanding.
\newblock \emph{arXiv preprint arXiv:2305.06355}, 2023{\natexlab{a}}.

\bibitem[Li et~al.(2024{\natexlab{b}})Li, Wang, He, Li, Wang, Liu, Wang, Xu, Chen, Luo, et~al.]{ref:mvbench}
Kunchang Li, Yali Wang, Yinan He, Yizhuo Li, Yi Wang, Yi Liu, Zun Wang, Jilan Xu, Guo Chen, Ping Luo, et~al.
\newblock Mvbench: A comprehensive multi-modal video understanding benchmark.
\newblock In \emph{Proceedings of the IEEE/CVF Conference on Computer Vision and Pattern Recognition}, pages 22195--22206, 2024{\natexlab{b}}.

\bibitem[Li et~al.(2023{\natexlab{b}})Li, Xie, Xie, Zhao, Zhang, Zheng, Zhao, and Zhang]{ref:momentdiff}
Pandeng Li, Chen-Wei Xie, Hongtao Xie, Liming Zhao, Lei Zhang, Yun Zheng, Deli Zhao, and Yongdong Zhang.
\newblock Momentdiff: Generative video moment retrieval from random to real.
\newblock In \emph{Advances in neural information processing systems}, pages 65948 -- 65966, 2023{\natexlab{b}}.

\bibitem[Li et~al.(2024{\natexlab{c}})Li, Xu, Zhang, Song, Cai, Qi, Zhou, Pan, Li, Vu, Huang, and Wang]{ref:groundinggpt}
Zhaowei Li, Qi Xu, Dong Zhang, Hang Song, Yiqing Cai, Qi Qi, Ran Zhou, Junting Pan, Zefeng Li, Van~Tu Vu, Zhida Huang, and Tao Wang.
\newblock Groundinggpt: Language enhanced multi-modal grounding model.
\newblock \emph{arXiv preprint arXiv:2401.06071}, 2024{\natexlab{c}}.

\bibitem[Lin et~al.(2023{\natexlab{a}})Lin, Ye, Zhu, Cui, Ning, Jin, and Yuan]{ref:videollava}
Bin Lin, Yang Ye, Bin Zhu, Jiaxi Cui, Munan Ning, Peng Jin, and Li Yuan.
\newblock Video-llava: Learning united visual representation by alignment before projection.
\newblock \emph{arXiv preprint arXiv:2311.10122}, 2023{\natexlab{a}}.

\bibitem[Lin et~al.(2023{\natexlab{b}})Lin, Tan, Hu, Jin, Ye, and Zheng]{ref:coll}
Zihang Lin, Chaolei Tan, Jian-Fang Hu, Zhi Jin, Tiancai Ye, and Wei-Shi Zheng.
\newblock Collaborative static and dynamic vision-language streams for spatio-temporal video grounding.
\newblock In \emph{Proceedings of the IEEE/CVF Conference on Computer Vision and Pattern Recognition}, pages 23100--23109, 2023{\natexlab{b}}.

\bibitem[Liu et~al.(2024{\natexlab{a}})Liu, Zeng, Ren, Li, Zhang, Yang, Jiang, Li, Yang, Su, Zhu, and Zhang]{ref:groundingdino}
Shilong Liu, Zhaoyang Zeng, Tianhe Ren, Feng Li, Hao Zhang, Jie Yang, Qing Jiang, Chunyuan Li, Jianwei Yang, Hang Su, Jun Zhu, and Lei Zhang.
\newblock Grounding dino: Marrying dino with grounded pre-training for open-set object detection.
\newblock In \emph{European Conference on Computer Vision}, pages 38--55, 2024{\natexlab{a}}.

\bibitem[Liu et~al.(2024{\natexlab{b}})Liu, Li, Liu, Wang, Ren, Li, Chen, Sun, and Hou]{ref:tempcompass}
Yuanxin Liu, Shicheng Li, Yi Liu, Yuxiang Wang, Shuhuai Ren, Lei Li, Sishuo Chen, Xu Sun, and Lu Hou.
\newblock Tempcompass: Do video llms really understand videos?
\newblock \emph{arXiv preprint arXiv:2403.00476}, 2024{\natexlab{b}}.

\bibitem[Loshchilov and Hutter(2017)]{ref:loshchilov2017decoupled}
Ilya Loshchilov and Frank Hutter.
\newblock Decoupled weight decay regularization.
\newblock \emph{arXiv preprint arXiv:1711.05101}, 2017.

\bibitem[Ma et~al.(2024)Ma, Jiang, Wu, Yuan, and Qi]{ref:groma}
Chuofan Ma, Yi Jiang, Jiannan Wu, Zehuan Yuan, and Xiaojuan Qi.
\newblock Groma: Localized visual tokenization for grounding multimodal large language models.
\newblock In \emph{European Conference on Computer Vision}, pages 417--435, 2024.

\bibitem[Maaz et~al.(2024{\natexlab{a}})Maaz, Rasheed, Khan, and Khan]{ref:vcgplus}
Muhammad Maaz, Hanoona Rasheed, Salman Khan, and Fahad Khan.
\newblock Videogpt+: Integrating image and video encoders for enhanced video understanding.
\newblock \emph{arXiv preprint arXiv:2406.09418}, 2024{\natexlab{a}}.

\bibitem[Maaz et~al.(2024{\natexlab{b}})Maaz, Rasheed, Khan, and Khan]{ref:videochatgpt}
Muhammad Maaz, Hanoona Rasheed, Salman Khan, and Fahad~Shahbaz Khan.
\newblock Video-chatgpt: Towards detailed video understanding via large vision and language models.
\newblock In \emph{Proceedings of the 62nd Annual Meeting of the Association for Computational Linguistics}, page 12585–12602, 2024{\natexlab{b}}.

\bibitem[Mangalam et~al.(2023)Mangalam, Akshulakov, and Malik]{ref:egoschema}
Karttikeya Mangalam, Raiymbek Akshulakov, and Jitendra Malik.
\newblock Egoschema: A diagnostic benchmark for very long-form video language understanding.
\newblock \emph{Advances in Neural Information Processing Systems}, 36:\penalty0 46212--46244, 2023.

\bibitem[Mao et~al.(2016)Mao, Huang, Toshev, Camburu, Yuille, and Murphy]{ref:refcocog}
Junhua Mao, Jonathan Huang, Alexander Toshev, Oana Camburu, Alan~L. Yuille, and Kevin Murphy.
\newblock Generation and comprehension of unambiguous object descriptions.
\newblock In \emph{Proceedings of the IEEE Conference on Computer Vision and Pattern Recognition}, pages 11--20, 2016.

\bibitem[Moon et~al.(2023)Moon, Hyun, Park, Park, and Heo]{ref:qddetr}
WonJun Moon, Sangeek Hyun, SangUk Park, Dongchan Park, and Jae-Pil Heo.
\newblock Query-dependent video representation for moment retrieval and highlight detection.
\newblock In \emph{Proceedings of the IEEE/CVF Conference on Computer Vision and Pattern Recognition}, pages 23023--23033, 2023.

\bibitem[Regneri et~al.(2013)Regneri, Rohrbach, Wetzel, Thater, Schiele, and Pinkal]{ref:tacos}
Michaela Regneri, Marcus Rohrbach, Dominikus Wetzel, Stefan Thater, Bernt Schiele, and Manfred Pinkal.
\newblock Grounding action descriptions in videos.
\newblock In \emph{Transactions of the Association for Computational Linguistics}, pages 25--36, 2013.

\bibitem[Ren et~al.(2024)Ren, Yao, Li, Sun, and Hou]{ref:timechat}
Shuhuai Ren, Linli Yao, Shicheng Li, Xu Sun, and Lu Hou.
\newblock Timechat: A time-sensitive multimodal large language model for long video understanding.
\newblock In \emph{Proceedings of the IEEE/CVF Conference on Computer Vision and Pattern Recognition}, pages 14313--14323, 2024.

\bibitem[Rezatofighi et~al.(2019)Rezatofighi, Tsoi, Gwak, Sadeghian, Reid, and Savarese]{ref:iou}
Hamid Rezatofighi, Nathan Tsoi, JunYoung Gwak, Amir Sadeghian, Ian Reid, and Silvio Savarese.
\newblock Generalized intersection over union: A metric and a loss for bounding box regression.
\newblock In \emph{Proceedings of the IEEE/CVF Conference on Computer Vision and Pattern Recognition}, pages 658--666, 2019.

\bibitem[Share(2024)]{ref:sharegemini}
Share.
\newblock Sharegemini: Scaling up video caption data for multimodal large language models.
\newblock \url{https://github.com/Share14/ShareGemini}, 2024.

\bibitem[Su et~al.(2021)Su, Yu, and Xu]{ref:stvgbert}
Rui Su, Qian Yu, and Dong Xu.
\newblock Stvgbert: A visual-linguistic transformer based framework for spatio-temporal video grounding.
\newblock In \emph{Proceedings of the IEEE/CVF International Conference on Computer Vision}, pages 1533--1542, 2021.

\bibitem[Tan et~al.(2021)Tan, Lin, Hu, Li, and Zheng]{tan2021augmented}
Chaolei Tan, Zihang Lin, Jian-Fang Hu, Xiang Li, and Wei-Shi Zheng.
\newblock Augmented 2d-tan: A two-stage approach for human-centric spatio-temporal video grounding.
\newblock \emph{arXiv preprint arXiv:2106.10634}, 2021.

\bibitem[Tang et~al.(2021)Tang, Liao, Liu, Li, Jin, Jiang, Yu, and Xu]{ref:hcstvg}
Zongheng Tang, Yue Liao, Si Liu, Guanbin Li, Xiaojie Jin, Hongxu Jiang, Qian Yu, and Dong Xu.
\newblock Human-centric spatio-temporal video grounding with visual transformers.
\newblock \emph{IEEE Transactions on Circuits and Systems for Video Technology}, 32\penalty0 (12):\penalty0 8238--8249, 2021.

\bibitem[Team et~al.(2024)Team, Georgiev, Lei, Burnell, Bai, Gulati, Tanzer, Vincent, Pan, Wang, et~al.]{ref:gemini}
Gemini Team, Petko Georgiev, Ving~Ian Lei, Ryan Burnell, Libin Bai, Anmol Gulati, Garrett Tanzer, Damien Vincent, Zhufeng Pan, Shibo Wang, et~al.
\newblock Gemini 1.5: Unlocking multimodal understanding across millions of tokens of context.
\newblock \emph{arXiv preprint arXiv:2403.05530}, 2024.

\bibitem[Wang et~al.(2024{\natexlab{a}})Wang, Xu, Cheng, Diao, Zhou, Cao, Wang, Ge, and Huang]{ref:grounded}
Haibo Wang, Zhiyang Xu, Yu Cheng, Shizhe Diao, Yufan Zhou, Yixin Cao, Qifan Wang, Weifeng Ge, and Lifu Huang.
\newblock Grounded-videollm: Sharpening fine-grained temporal grounding in video large language models.
\newblock \emph{arXiv preprint arXiv:2410.03290}, 2024{\natexlab{a}}.

\bibitem[Wang et~al.(2024{\natexlab{b}})Wang, Ye, Wang, Nie, and Huang]{ref:elysium}
Han Wang, Yongjie Ye, Yanjie Wang, Yuxiang Nie, and Can Huang.
\newblock Elysium: Exploring object-level perception in videos via mllm.
\newblock In \emph{European Conference on Computer Vision}, pages 166--185, 2024{\natexlab{b}}.

\bibitem[Wang et~al.(2023)Wang, He, Li, Li, Yu, Ma, Li, Chen, Chen, Wang, He, Luo, Liu, Wang, Wang, and Qiao]{ref:intervid}
Yi Wang, Yinan He, Yizhuo Li, Kunchang Li, Jiashuo Yu, Xin Ma, Xinhao Li, Guo Chen, Xinyuan Chen, Yaohui Wang, Conghui He, Ping Luo, Ziwei Liu, Yali Wang, Limin Wang, and Yu Qiao.
\newblock Internvid: A large-scale video-text dataset for multimodal understanding and generation.
\newblock In \emph{arXiv preprint arXiv:2307.06942}, 2023.

\bibitem[Wang et~al.(2024{\natexlab{c}})Wang, Meng, Liang, Wang, Liu, and Zhao]{ref:hawkeye}
Yueqian Wang, Xiaojun Meng, Jianxin Liang, Yuxuan Wang, Qun Liu, and Dongyan Zhao.
\newblock Hawkeye: Training video-text llms for grounding text in videos.
\newblock \emph{arXiv preprint arXiv:2403.10228}, 2024{\natexlab{c}}.

\bibitem[Wang et~al.(2022)Wang, Wang, Wu, Li, and Wu]{wang2022negative}
Zhenzhi Wang, Limin Wang, Tao Wu, Tianhao Li, and Gangshan Wu.
\newblock Negative sample matters: A renaissance of metric learning for temporal grounding.
\newblock In \emph{Proceedings of the AAAI Conference on Artificial Intelligence}, pages 2613--2623, 2022.

\bibitem[Xiao et~al.(2021)Xiao, Shang, Yao, and Chua]{ref:nextqa}
Junbin Xiao, Xindi Shang, Angela Yao, and Tat-Seng Chua.
\newblock Next-qa: Next phase of question-answering to explaining temporal actions.
\newblock In \emph{Proceedings of the IEEE/CVF conference on computer vision and pattern recognition}, pages 9777--9786, 2021.

\bibitem[Yang et~al.(2022)Yang, Miech, Sivic, Laptev, and Schmid]{ref:tubedetr}
Antoine Yang, Antoine Miech, Josef Sivic, Ivan Laptev, and Cordelia Schmid.
\newblock Tubedetr: Spatio-temporal video grounding with transformers.
\newblock In \emph{Proceedings of the IEEE/CVF Conference on Computer Vision and Pattern Recognition}, pages 16442--16453, 2022.

\bibitem[Yang et~al.(2024{\natexlab{a}})Yang, Yang, Hui, Zheng, Yu, Zhou, Li, Li, Liu, Huang, et~al.]{yang2024qwen2}
An Yang, Baosong Yang, Binyuan Hui, Bo Zheng, Bowen Yu, Chang Zhou, Chengpeng Li, Chengyuan Li, Dayiheng Liu, Fei Huang, et~al.
\newblock Qwen2 technical report.
\newblock \emph{arXiv preprint arXiv:2407.10671}, 2024{\natexlab{a}}.

\bibitem[Yang et~al.(2024{\natexlab{b}})Yang, Yang, Zhang, Hui, Zheng, Yu, Li, Liu, Huang, Wei, et~al.]{ref:qwen2.5}
An Yang, Baosong Yang, Beichen Zhang, Binyuan Hui, Bo Zheng, Bowen Yu, Chengyuan Li, Dayiheng Liu, Fei Huang, Haoran Wei, et~al.
\newblock Qwen2.5 technical report.
\newblock \emph{arXiv preprint arXiv:2412.15115}, 2024{\natexlab{b}}.

\bibitem[Yi et~al.(2020)Yi, Gan, Li, Kohli, Wu, Torralba, and B.Tenenbaum]{ref:clevrer}
Kexin Yi, Chuang Gan, Yunzhu Li, Pushmeet Kohli, Jiajun Wu, Antonio Torralba, and Joshua B.Tenenbaum.
\newblock Clevrer: Collision events for video representation and reasoning.
\newblock In \emph{International Conference on Learning Representations}, 2020.

\bibitem[You et~al.(2023)You, Zhang, Gan, Du, Zhang, Wang, Cao, Chang, and Yang]{ref:fer}
Haoxuan You, Haotian Zhang, Zhe Gan, Xianzhi Du, Bowen Zhang, Zirui Wang, Liangliang Cao, Shih-Fu Chang, and Yinfei Yang.
\newblock Ferret: Refer and ground anything anywhere at any granularity.
\newblock \emph{arXiv preprint arXiv:2310.07704}, 2023.

\bibitem[Yu et~al.(2021)Yu, Wang, Hu, Luo, and Li]{yu20212rd}
Yi Yu, Xinying Wang, Wei Hu, Xun Luo, and Cheng Li.
\newblock 2rd place solutions in the hc-stvg track of person in context challenge 2021.
\newblock \emph{arXiv preprint arXiv:2106.07166}, 3\penalty0 (7), 2021.

\bibitem[Yu et~al.(2019)Yu, Xu, Yu, Yu, Zhao, Zhuang, and Tao]{ref:activitynetqa}
Zhou Yu, Dejing Xu, Jun Yu, Ting Yu, Zhou Zhao, Yueting Zhuang, and Dacheng Tao.
\newblock Activitynet-qa: A dataset for understanding complex web videos via question answering.
\newblock In \emph{Proceedings of the AAAI Conference on Artificial Intelligence}, pages 9127--9134, 2019.

\bibitem[Zhai et~al.(2023)Zhai, Mustafa, Kolesnikov, and Beyer]{ref:siglip}
Xiaohua Zhai, Basil Mustafa, Alexander Kolesnikov, and Lucas Beyer.
\newblock Sigmoid loss for language image pre-training.
\newblock In \emph{Proceedings of the IEEE/CVF international conference on computer vision}, pages 11975--11986, 2023.

\bibitem[Zhang et~al.(2023)Zhang, Li, and Bing]{ref:videollama}
Hang Zhang, Xin Li, and Lidong Bing.
\newblock Video-llama: An instruction-tuned audio-visual language model for video understanding.
\newblock \emph{arXiv preprint arXiv:2306.02858}, 2023.

\bibitem[Zhang et~al.(2024{\natexlab{a}})Zhang, Li, Li, Ren, Zou, Liu, Huang, Gao, Leizhang, Li, and Yang]{ref:llavagrounding}
Hao Zhang, Hongyang Li, Feng Li, Tianhe Ren, Xueyan Zou, Shilong Liu, Shijia Huang, Jianfeng Gao, Leizhang, Chunyuan Li, and Jainwei Yang.
\newblock Llava-grounding: Grounded visual chat with large multimodal models.
\newblock In \emph{European Conference on Computer Vision}, pages 19--35, 2024{\natexlab{a}}.

\bibitem[Zhang et~al.(2024{\natexlab{b}})Zhang, Li, Zhang, Pu, Cahyono, Hu, Liu, Zhang, Yang, Li, et~al.]{zhang2024lmms}
Kaichen Zhang, Bo Li, Peiyuan Zhang, Fanyi Pu, Joshua~Adrian Cahyono, Kairui Hu, Shuai Liu, Yuanhan Zhang, Jingkang Yang, Chunyuan Li, et~al.
\newblock Lmms-eval: Reality check on the evaluation of large multimodal models.
\newblock \emph{arXiv preprint arXiv:2407.12772}, 2024{\natexlab{b}}.

\bibitem[Zhang et~al.(2024{\natexlab{c}})Zhang, Zhang, Li, Zeng, Yang, Zhang, Wang, Tan, Li, and Liu]{ref:longva}
Peiyuan Zhang, Kaichen Zhang, Bo Li, Guangtao Zeng, Jingkang Yang, Yuanhan Zhang, Ziyue Wang, Haoran Tan, Chunyuan Li, and Ziwei Liu.
\newblock Long context transfer from language to vision.
\newblock \emph{arXiv preprint arXiv:2406.16852}, 2024{\natexlab{c}}.

\bibitem[Zhang et~al.(2024{\natexlab{d}})Zhang, Wu, Li, Li, Ma, Liu, and Li]{ref:llavavideo}
Yuanhan Zhang, Jinming Wu, Wei Li, Bo Li, Zejun Ma, Ziwei Liu, and Chunyuan Li.
\newblock Video instruction tuning with synthetic data.
\newblock \emph{arXiv preprint arXiv:2410.02713}, 2024{\natexlab{d}}.

\bibitem[Zhang et~al.(2020)Zhang, Zhao, Zhao, Wang, Liu, and Gao]{ref:vidstvg}
Zhu Zhang, Zhou Zhao, Yang Zhao, Qi Wang, Huasheng Liu, and Lianli Gao.
\newblock Where does it exist: Spatio-temporal video grounding for multi-form sentences.
\newblock In \emph{Proceedings of the Twenty-Ninth International Joint Conference on Artificial Intelligence}, pages 10668--10677, 2020.

\bibitem[Zhang et~al.(2021)Zhang, Zhao, Lin, Huai, and Yuan]{ref:object}
Zhu Zhang, Zhou Zhao, Zhijie Lin, Baoxing Huai, and Jing Yuan.
\newblock Object-aware multi-branch relation networks for spatio-temporal video grounding.
\newblock In \emph{Proceedings of the Twenty-Ninth International Joint Conference on Artificial Intelligence}, pages 1069--1075, 2021.

\end{thebibliography}
}


\end{document}